\documentclass[conference,letterpaper,twoside,final,10pt]{ieeeconf}
\usepackage[left=1.91cm,right=1.91cm,top=1.91cm,bottom=1.91cm]{geometry}

\usepackage{graphicx}
\usepackage{amssymb}
\usepackage{amsmath}
\usepackage{bm}
\usepackage{cite}
\usepackage{comment}
\usepackage{color}
\usepackage{url}
\usepackage{alltt}
\usepackage{cleveref}
\usepackage{tikz}
\usepackage{pgfplots}
\usepackage{pgf-umlsd}
\usepackage{cprotect}
\usepackage{multirow}
\usepackage{algorithm}
\usepackage[noend]{algpseudocode}
\usepackage{bigfoot}
\usepackage{paralist}
\usepackage{tabularx}
\usepackage[whole]{bxcjkjatype}
\usepackage{ifthen}
\usepackage{threeparttable}
\usepackage[hang,small,bf]{caption}
\usepackage[subrefformat=parens]{subcaption}
\usepackage[super]{nth}
\usepackage{colortbl}
\usepackage{scalefnt}

\pgfplotsset{compat=1.14}
\IEEEoverridecommandlockouts

\title{\LARGE \bf
Deep Gated Multi-modal Learning:\\
In-hand Object Pose Changes Estimation using Tactile and Image Data
}

\author{
    Tomoki Anzai$^{*\dagger}$,
    Kuniyuki Takahashi$^{*\ddagger}$
    \thanks{$^{*}$ The starred authors have contributed equally.}
    \thanks{$^{\dagger}$ T. Anzai is associated with The University of Tokyo. This work is an achievement during internship and part-time job at Preferred Networks. \ anzai@jsk.imi.i.u-tokyo.ac.jp}
    \thanks{$^{\ddagger}$K. Takahashi is associated with Preferred Networks, Inc. \newline
    takahashi@preferred.jp}
}

\begin{document}

\maketitle
\thispagestyle{empty}

\begin{abstract}
For in-hand manipulation, estimation of the object pose inside the hand is one of the important functions to manipulate objects to the target pose.
Since in-hand manipulation tends to cause occlusions by the hand or the object itself, image information only is not sufficient for in-hand object pose estimation.
Multiple modalities can be used in this case, the advantage is that other modalities can compensate for occlusion, noise, and sensor malfunctions. 
Even though deciding the utilization rate of a modality (referred to as \emph{reliability value}) corresponding to the situations is important, the manual design of such models is difficult, especially for various situations.
In this paper, we propose \emph{deep gated multi-modal learning}, which self-determines the \emph{reliability value} of each modality through end-to-end deep learning.
For the experiments, an RGB camera and a GelSight tactile sensor were attached to the parallel gripper of the Sawyer robot, and the object pose changes were estimated during grasping.
A total of 15 objects were used in the experiments.
In the proposed model, the \emph{reliability values} of the modalities were determined according to the noise level and failure of each modality, and it was confirmed that the pose change was estimated even for unknown objects.
\footnote{An accompanying video is available at the following link:\\ \url{https://youtu.be/2OPQxQIcSxY}}
\end{abstract}

\section{Introduction}
\label{sec:introduction}
Robots are expected to not only work in factories, but also in home situations where they are expected to grasp objects in various environments.
Reaching the target grasp pose in a direct motion is difficult when there are obstacles in the environment.
In such situations, the robots need to grasp the object once and then place it on a surface such as a desk to grasp it again, in order to achieve the target pose.
Another way is to rearrange the object inside the hand, also known as in-hand manipulation.
Though commonly seen as the easier solution, placing and re-grasping an object generally takes more time and steps compared to in-hand manipulation.
Furthermore, surfaces to place the object on do not always exist in environments, therefore this method can not always be used.
For robot manipulation tasks, particularly in-hand manipulation, object pose estimation within a robotic hand is one of the important functions to manipulate objects to the target pose accurately~\cite{yousef2011tactile, nikhil2018inhand}.

\begin{figure}[tb]
    \centering
    \includegraphics[width=1.0\columnwidth]{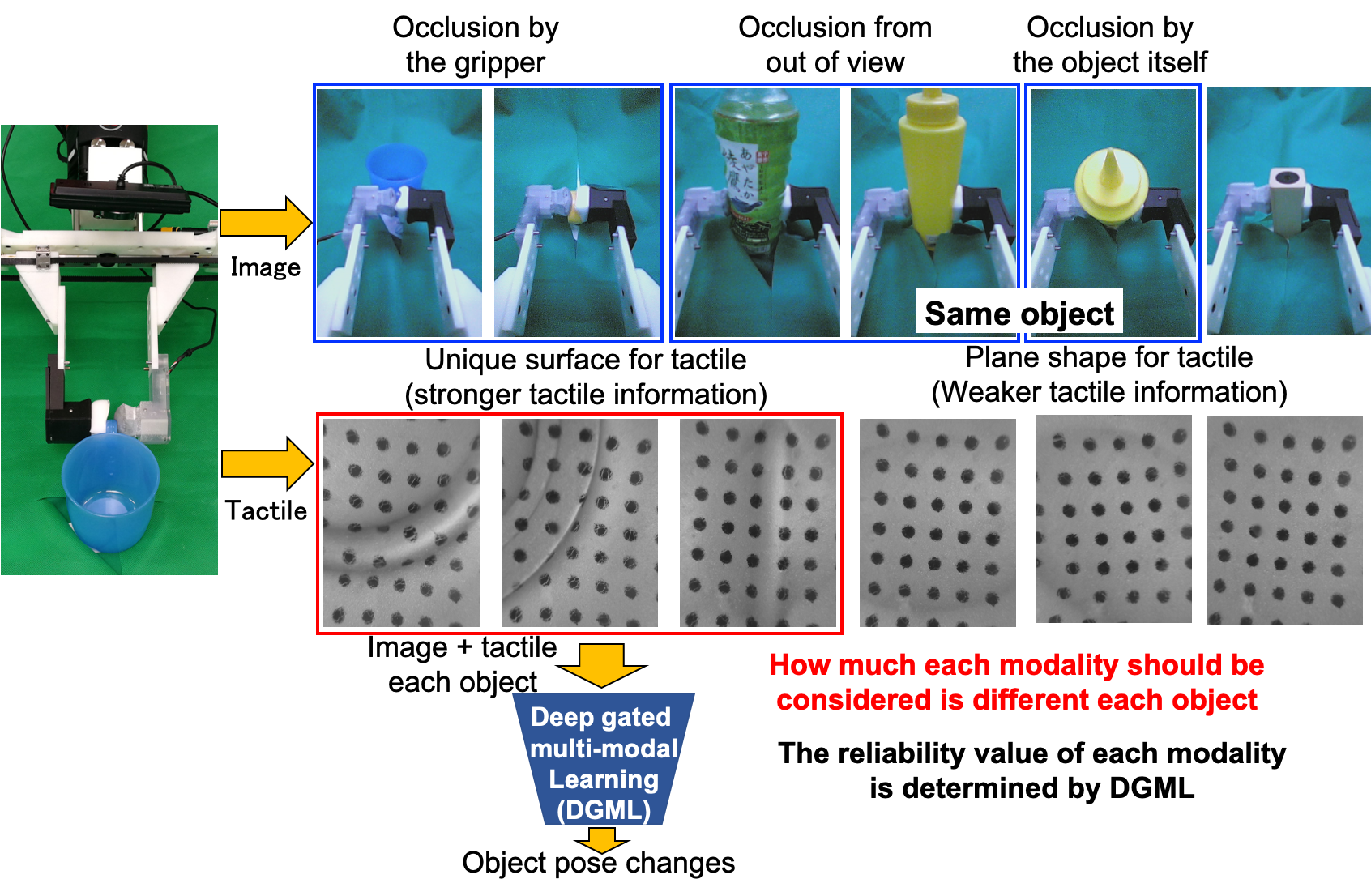}
    \caption{An illustration of in-hand object pose estimation with image and tactile data through our proposed network, deep gated multi-modal learning, which can decide the \emph{reliability value} for each modality.}
    \label{fig:representation_research}
\end{figure}

Although many researchers have developed object pose estimation based on image information only, images alone are not sufficient for in-hand manipulation, where the object is likely occluded from the camera by the object or the hand itself (e.g. the gripper hides the object).
Furthermore, large objects can go outside the field of vision of the camera or hide parts of themselves as depicted in Fig.~\ref{fig:representation_research}.
One of the approaches to address these challenges is to combine multiple sensor modalities.
One of the advantages of using multiple modalities is that other modalities can compensate and provide information from different perspectives, even in the case of occlusions, noise, or malfunctions in some of the other modalities.
In such situations, it is necessary to predict how much each modality should be considered.
For the remainder of this article, we refer to the utilization ratio of each modality as its \textit{reliability value}.
However, it is difficult to manually design a model that can decide a given modality's \emph{reliability value}, especially if the model has to deal with a wide variety of environments and situations.

\begin{figure*}[t]
    \vspace{2mm}
    \centering
    \includegraphics[width=1.90\columnwidth]{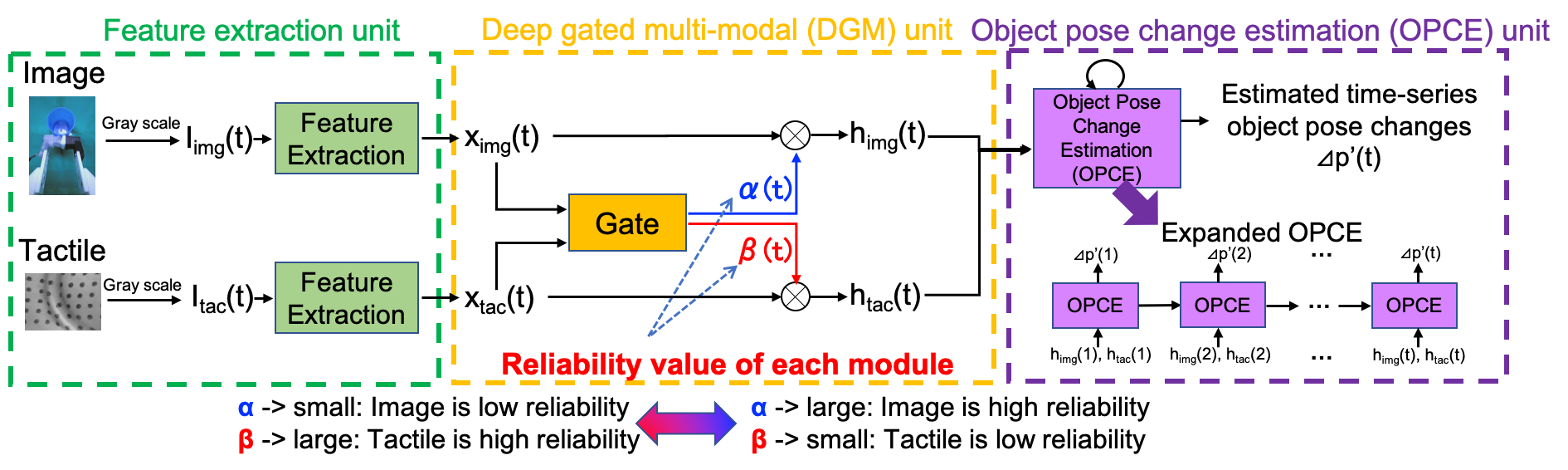}
    \caption{The proposed network architecture for deep gated multi-modal learning.
    Inputs are sequences of image and tactile data, and the output is a time-series of object pose changes.
    The gate values ($\alpha$ and $\beta$) represent the reliability of each module, and the values are acquired by the network itself.
    After training, a low gate value results in a low \emph{reliability value}, whereas a high gate value results in a high \emph{reliability value}.
    }
    \label{fig:network_model}
\end{figure*}

In this paper, we propose a method that we call \emph{deep gated multi-modal learning} (DGML), which uses end-to-end deep learning to predict and determine the \emph{reliability value} of each modality by DGML itself (See Fig.~\ref{fig:representation_research} and Fig.~\ref{fig:network_model}).
By virtue of end-to-end deep learning, this method is capable of generalizing to unknown objects without assuming a known 3D model when doing in-hand pose estimation.

The rest of this paper is organized as follows.
The contribution is explained in~\Cref{sec:contribution} and works related to this paper are described in~\Cref{sec:related works}, while our proposed method is detailed in ~\Cref{sec:proposed_method}.
\Cref{sec:experimental setup} outlines our experiment setup, and evaluation settings with the results are presented in~\Cref{sec:result}.
Finally, future work and conclusions are explained in~\Cref{sec:conclusion}.

\section{Contributions}
\label{sec:contribution}
The target of our method was to estimate the object pose \textbf{changes} inside the hand using image and tactile sensors as robustly as possible, despite occlusions, noise, sensor malfunctions and other possible obstructions.
Note that we track \textbf{changes} (plural) since we estimate a time-series of how the object pose evolves per time step.
What should also be noted is that our method estimates in-hand \textbf{relative object pose changes} during grasping instead of the absolute object pose with respect to the robot base, meaning that we estimate how much the object moves within the hand after the robot first makes contact with the object (when it has grasped).
This new method is able to dynamically determine the \emph{reliability value} and scale each modality's contribution appropriately instead of focusing on only one modality or the other.

The main contributions of this article are as follows:
\begin{itemize}
    \item Proposed a new approach in which the network itself determines the reliability of each modality dynamically and uses that \emph{reliability value} to scale each modality's contribution.
    \item Proposed a new approach in which end-to-end learning combines image and tactile data without assuming a 3D model to estimate the pose changes of unseen objects which is robust against occlusions, noise, and sensor malfunctions.
    \item Investigated details of noise, malfunction, and occlusion behavior in sensor information unique to the robot field.
\end{itemize}

\section{Related Works}
\label{sec:related works}
\subsection{Object Pose Estimation with Depth and Image Data}
\label{sec:related_works_pose_image}
Object pose estimation is a well-studied problem in computer vision and is important for robotic tasks.
Many researchers have particularly been developing methods using depth data (point cloud) or RGB-D data~\cite{choi2012voting, choi20123d, aldoma2012tutorial}.
Classical approaches with depth data are mainly based on point cloud matching methods such as iterative closest point (ICP)~\cite{ICP}.
These methods can achieve high accuracy, but since these methods require 3D models of the objects in advance, they cannot be used for unknown objects.

Recently, deep learning has become an active research area and especially the computer vision field has achieved success. 
Image based object pose estimation methods through combination of deep learning with model-based approaches have been studied~\cite{krull2015learning, xiang2017posecnn, Wang2019DenseFusion}.
The convergence to the final result of ICP heavily relies on the choice of the initial position, but the convergence error can be suppressed by giving a proper initial position through deep-learning methods.
However, these methods still require a 3D model, thus adapting to unknown objects remains challenging.
Some state-of-the-arts pose estimation methods that do not require 3D models have been realized through deep learning~\cite{schwarz2015rgbd, hodan2018bop, hu2019segmentation, andrychowicz2020learning}.
\subsection{Tactile-based Object Pose Estimation}
\label{sec:related_works_tactile}
As described in~\Cref{sec:related_works_pose_image}, most of the existing pose estimation methods use images and depth information only.
During in-hand object manipulation, the object is occluded from the camera and depth sensor by the hand or the object itself (See Fig.~\ref{fig:representation_research} as examples of occlusions).

Tactile sensors are gaining attention since they can observe the contact state even with occlusions~\cite{tomo2016uskin, tomo2018uskin, yuan2017gelsight, dong2017improved}.
The majority of these sensors fall in either of the following two categories:
\begin{enumerate}
    \item Single-axis multi-touch sensors, which can only sense normal force~\cite{ohmura2006conformable,iwata2009design,mittendorfer2011humanoid,fishel2012sensing}
    \item Three-axis single-touch sensors~\cite{paulino2017low}
\end{enumerate}
Two of the few exceptions are the uSkin~\cite{tomo2018uskin} and the GelSight~\cite{johnson2009retrographic, dong2017improved} which are multi-touch sensors that can measure shear force as well as normal force.
The commercialized uSkin sensor~\cite{tomo2018uskin} utilizes embedded magnets inside a silicone rubber and measures the deformation of the silicone during contact by monitoring changes in the magnetic fields.
Using this method, it is able to measure both normal as well as shear force per sensor unit. 
The prototype supports 16 different points of contact~\cite{tomo2016uskin}.
Instead of magnets, GelSight~\cite{yuan2017gelsight, Calandra2017} is an optical-based tactile sensor, which uses a camera to capture and measure the deformation of the attached elastomer during contact with a surface.

There are some works using tactile sensors for pose estimation such as~\cite{bimbo2015global}, but a 3D model of the object is required a priori.
It is challenging to estimate the pose of an object using only the tactile information of the grasped part since there is little information from only a single grasp.
Therefore, some work has been done in which tactile and image information are combined~\cite{gao2016deep, Calandra2017, Bimbo2012}, and the effectiveness of this fusion was demonstrated in object grasping tasks.
Some of the works performed object pose estimation through model based approaches using known 3D models~\cite{honda1998real, hebert2011fusion, Bimbo2016}.
In these studies, a 3D model of the object is often required to estimate absolute object pose because it is assumed that the pose of the object is unknown when the object is grasped.
However, we expect that even the methods that work on unknown objects described in~\ref{sec:related_works_pose_image} can be improved by using additional tactile information for dealing with occlusions.
The object pose can namely be estimated \textbf{before} the robot grasps an object, through methods without a 3D model of the object such as~\cite{schwarz2015rgbd, hodan2018bop, hu2019segmentation} earlier mentioned in~\ref{sec:related_works_pose_image}.
If we then keep track of the object pose changes using a tactile sensor \textbf{after} the robot first makes contact (when it has grasped the object and thus occludes it), we will be able to track the absolute object pose without using a given 3D model.
In addition, our hypothesis is that tactile sensors improve the object pose estimation accuracy compared to using images only (Details of our experiment results to test this are described in Table~\ref{tab:accuracy} in section~\ref{sec:inference}).

\subsection{Attention Based Learning}
A variety of attention methods has been studied.
Regarding the division of attention to each modality, the majority of these methods fall in either of the following three categories, except for our proposed method:

1) Equal attention to all modalities:
Each modality is fed into their respective networks to extract features, and the obtained features are simply combined to estimate information such as grasping point and motion.
For example, the combination of image and tactile data described in~\Cref{sec:related_works_tactile}, motion and image~\cite{takahashi2017dynamic}, force and image~\cite{lee2019making}, language and image~\cite{hatori2018interactive}, and, image and sound~\cite{noda2014multimodal}.

2) Attention within modalities:
Only the important parts of the given modalities are used.
Within each modality, the part to focus on is extracted by the network itself.
For example, in the case of images, the pixel of interest is used instead of the whole image~\cite{kim2018robust, hu2018squeeze}.
This case does not necessarily use multiple modalities as it can also be done when using a single modality.

3) Cancellation of irrelevant modalities:
Only the important modalities are used while other modalities are ignored.
The network decides which modality should be used from all modalities.
For example, choosing to utilize language or images according to the situation~\cite{arevalo2017gated}.

The new approach we developed does not fit in these categories, since it determines the contribution ratio of each modality dynamically and uses that \emph{reliability value} to scale each modality's contribution.

\section{Deep Gated Multi-modal learning}
\label{sec:proposed_method}
We propose a method that we call \emph{deep gated multi-modal learning} (DGML) for in-hand object pose changes estimation with image and tactile information based on end-to-end deep learning.
\textbf{The DGML estimates a time-series of relative in-hand object pose changes during grasping instead of the absolute object pose.}
Figure~\ref{fig:network_model} shows the concept of the proposed network model.
The main concept of DGML is that the network itself dynamically decides how much each modality it should be relied on, in other words to decide the \textit{reliability value} per modality.
We aimed to design a network with a structure that is as simple as possible, but still sufficient to show the effectiveness of the deep gated multi-modal unit.
Increased complexity of the network architecture will most likely improve the accuracy, but the concept of \emph{reliability values} can be used in the same way.
The details of the network structure will be described in section~\ref{sec:Network_Design}.

DGML is composed of three components for in-hand pose estimation:
\begin{itemize}
    \item Feature extraction unit to extract features from image and tactile data
    \item Deep gated multi-modal unit for calculation and application of a \emph{reliability value} for each modality
    \item Object pose changes estimation unit to estimate object pose changes using time-series input information
\end{itemize}
For training and inference of the network, a sequence of image and tactile data are the input, and the output is a sequence of object pose changes.
A training dataset $D = \{(input_{1},output_{1}),...,(input_{T},output_{T})\}_{n}$ consists of $n$ sequence with $T$ steps. 

As a feature extraction unit, convolutional neural networks (CNNs) are used to calculate image features $x_{img}(t)$ and tactile features $x_{tac}(t)$ at step $t$ from image input $I_{img}(t)$ and tactile input $I_{tac}(t)$, respectively.

Then, the \emph{reliability value} $\alpha$ for image in step $t$ is given from $Gate$ as:
\begin{eqnarray}
    \alpha(t)  &=& Gate(x_{img}(t), x_{tac}(t)) \nonumber\\
               &=& Sigmoid(FC_{img}(x_{img}(t)) + \nonumber\\
               & & \qquad \qquad \qquad \qquad FC_{tac}(x_{tac}(t)))
    \label{eq:gate}
\end{eqnarray}
where $FC_{img}$ and $FC_{tac}$ are fully connected (FC) layers to reduce the number of dimensions, and $Sigmoid$ is used as activation function for the gate.
The \emph{reliability values} $\alpha(t)$ for image and $\beta(t)$ for tactile data are conditioned to sum up to 1:
\begin{equation}
    \alpha(t) + \beta(t) = 1
    \label{eq:condition}
\end{equation}
$\alpha(t)$ and $\beta(t)$ are one-dimensional scalars.
The gate determines each modality's \emph{reliability values} from all modalities' information through equations~(\ref{eq:gate}) and~(\ref{eq:condition}).

The extracted feature vectors for each modality, $x_{img}(t)$ and $x_{tac}(t)$, are then scaled by multiplying them with their respective \emph{reliability values}, giving us the scaled feature vectors $h_{img}(t)$ and $h_{tac}(t)$ for time step $t$:
\begin{eqnarray}
    h_{img}(t) &=& \alpha(t) \cdot x_{img}(t) \nonumber\\
    h_{tac}(t) &=& \beta(t) \cdot x_{tac}(t)
    \label{eq:multiplied}
\end{eqnarray}
Since a sigmoid function is used and the total sum of \emph{reliability values} is 1, \textbf{the \emph{reliability values} are continuous values from 0 to 1, instead of being only 0 or 1}.
The lower the reliability, the smaller the contribution to the output.
On the other hand, the greater the reliability, the greater the output contribution.
In addition, if the \emph{reliability value} becomes 0, the modality is completely ignored.
Therefore, it is not necessary to manually determine which sensor to enable/disable in advance, because the network will automatically ignore them if they are not helpful.
Note that \textbf{the \emph{reliability values} are not absolute, but relative}.
We chose to use relative values to learn the correlation between multiple modalities.
This however means that the method simply assigns a higher \emph{reliability value} to a given modality, if said modality performs better than any other modality.
Thus, all modalities could be noisy or under-performing, but our method will still assign \emph{reliability values} by comparing the modalities' performances with each other.
The downside of this is that the robot will continue to operate even if none of the sensor modalities are performing well.
Rather, it tries to make the best out of the situation and attempts to focus on only the good modalities of the data it has.
Detecting when the robot should give up trying or improving the quality of the data however is out of scope of our work.

As the object pose changes estimation unit, long short-term memory (LSTM) with its output connected to FC is used.
Input of the LSTM is a sequence of $h_{img}(t)$ and $h_{tac}(t)$, whereas the output of FC is $O^{\prime}(t)=\Delta{p}^{\prime}(t) {\in} {\bm{R}}^{d}$, which is the estimated sequence of object pose changes from the LSTM.

Since our proposed method works without 3D models given in advance, \textbf{we estimate the relative object pose (object pose changes) during grasping instead of the absolute pose}.
Thus, object pose change $\Delta{p}(t)$ is the current object pose at step $t$ with respect to the initial object pose at step $1$ when the robot first grasps the object and makes contact.
When $t=1$,  $O(1)= \Delta{p}(1) = \textbf{0}$.


The loss function $L$ is minimized as follows:
\begin{equation}
    \min_{\xi}\frac{1}{mT}\sum_{i=1}^{m}\sum_{t=1}^{T} L(O(t), O^{\prime}(t))
    \label{eq:cost_eq}
\end{equation}
where $\xi$ are the parameters to be trained, $m(\leqq n)$ is the number of sequences for mini-batch training, and $O(t) {\in} {\bm{R}}^{d}$ is the teaching signal.

There is no teaching signal for the \emph{reliability values} $\alpha$ and $\beta$ because these are calculated by the network itself to minimize the output error by equation (\ref{eq:cost_eq}).
Then, the \emph{reliability values} are multiplied by each modality in equation (\ref{eq:multiplied}).


\section{Experimental Setup}
\label{sec:experimental setup}
The purpose of the experiments is to verify DGML in situations with occlusion, noise, and sensor malfunctions.

We note that our values of the hyper-parameters provided in this section are tuned by random search.
\subsection{Hardware Setup}
\subsubsection{Tactile sensor}
The GelSight tactile sensor we duplicate from article~\cite{yuan2017gelsight, Calandra2017} is an optical-based tactile sensor, which captures a $640\times480$ image by a camera, which can then be used to calculate the 3D model and the applied normal and shear force in the x, y, and z axes in Newton.
However, instead of using the 3D model and the force in Newton we directly use the captured raw image from the GelSight (See Fig.~\ref{fig:representation_research}).
Because the silicone layer from the sensor's container tears when an excess amount of force is applied, we applied baby powder on this layer to reduce friction between the surface and the grasped object.
The reason why we chose GelSight is that it is a multi-touch sensor that can measure both normal and shear forces as opposed to other types of tactile sensors described in section~\ref{sec:related_works_tactile}.
It is also relatively cheap and simple to reproduce.

\begin{figure}
    \vspace{2mm}
    \centering
    \includegraphics[width=0.80\columnwidth]{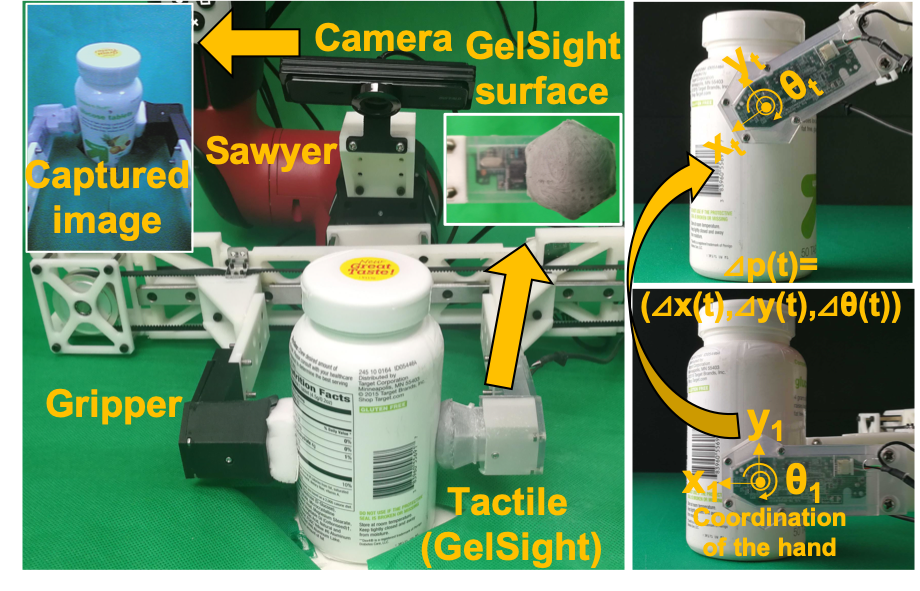}
    \caption{Setup used in our experiments.
            Custom printed end-effector with both a tactile skin sensor and a web camera.
            The Sawyer robot let the gripper move to the minus x-axis, y-axis direction and rotation of yaw.}
    \label{fig:hardware}
\end{figure}

\begin{figure}[tb]
    \centering
    \includegraphics[width=0.70\columnwidth]{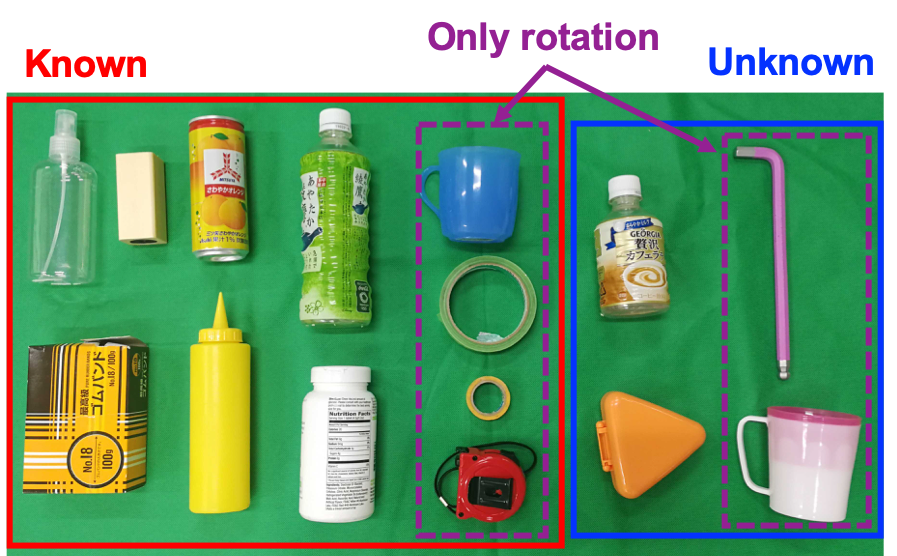}
    \caption{Trained objects (red) and unknown objects (blue)}
    \label{fig:objects}
\end{figure}
\subsubsection{Gripper}
We developed a gripper to grasp objects shown in Fig.~\ref{fig:hardware}.
This gripper is a parallel gripper which has two fingers driven by a servo motor (Dynamixel XM430-W350-R).
A GelSight tactile sensor is attached to one fingertip, and the other fingertip has a sponge.
A camera (BUFFALO BSW200MBK) is mounted at the center of the gripper.

\subsubsection{Sawyer}
To perform our experiments, we use a Sawyer 7-DOF robotic arm with the gripper as end-effector (See Fig.~\ref{fig:hardware}).
The Sawyer, GelSight sensor, gripper, and camera are connected to a PC running Ubuntu 16.04 with ROS Kinetic.
\begin{table}[tb]
    \vspace{2mm}
    \centering
    \begin{threeparttable}
    \caption{Network Design\tnote{1}}
    \label{tab:networkdesign}
    \begingroup
    \scalefont{0.90}
    \begin{tabular}{|c|c|c||c|c|c|c|c|}
    \hline
    & & Layer & In & Out & \begin{tabular}{c}Filter\\size\end{tabular} & \begin{tabular}{c}Activation\\function\end{tabular} \\
    \hline\hline
     \parbox[t]{2mm}{\multirow{18}{*}{\rotatebox[origin=c]{90}{Feature Extraction Unit}}}
     &\parbox[t]{2mm}{\multirow{9}{*}{\rotatebox[origin=c]{90}{Image feature extraction}}}
     		&  \nth{1} conv. & 1  & 32 & (3,3) & ReLu \\
     		&& \nth{2} conv. & 32  & 32 & (3,3) & ReLu \\
     		&& Ave. pooling & 32 & 32 & (4,4) & - \\
     		&& \nth{3} conv. & 32 & 32 & (3,3) & ReLu \\
     		&& \nth{4} conv. & 32 & 32 & (3,3) & ReLu \\
     		&& Ave. pooling & 32 & 32 & (2,2) & - \\
     		&& \nth{5} conv. & 32 & 32 & (3,3) & ReLu \\
     		&& \nth{6} conv. & 32 & 32 & (3,3) & ReLu \\
     		&& Ave. pooling & 32 & 32 & (2,2) & - \\
    \cline{2-7}
     &\parbox[t]{2mm}{\multirow{9}{*}{\rotatebox[origin=c]{90}{Tactile feature extraction}}}
     		&  \nth{1} conv. & 1  & 32 & (3,3) & ReLu \\
     		&& \nth{2} conv. & 32 & 32 & (3,3) & ReLu \\
     		&& Ave. pooling & 32 & 32 & (4,4) & - \\
     		&& \nth{3} conv. & 32 & 32 & (3,3) & ReLu \\
     		&& \nth{4} conv. & 32 & 32 & (3,3) & ReLu \\
     		&& Ave. pooling & 32 & 32 & (2,2) & - \\
     		&& \nth{5} conv. & 32 & 32 & (3,3) & ReLu \\
     		&& \nth{6} conv. & 32 & 32 & (3,3) & ReLu \\
     		&& Ave. pooling & 32 & 32 & (2,2) & - \\
    \hline
     \parbox[t]{2mm}{\multirow{6}{*}{\rotatebox[origin=c]{90}{DGM Unit}}}
     &\parbox[t]{2mm}{\multirow{6}{*}{\rotatebox[origin=c]{90}{Gate}}}
     		& \nth{1} $FC_{img}$   & 1120          & 1 & - & - \\
     		&&                      & {\tiny ($x_{img}(t)$)} &  &  &  \\
     		&& \nth{1} $FC_{tac}$   & 2240          & 1 & - & - \\
     		&&                      & {\tiny ($x_{tac}(t)$)} &  &  &  \\
     		&& \nth{2} Gate         & 2             & 1 & - & sigmoid \\
     		&&                      &               & {\tiny ($\alpha(t)$)}  &   &   \\
    \hline
     \parbox[t]{2mm}{\multirow{5}{*}{\rotatebox[origin=c]{90}{OPCE Unit}}}
     &\parbox[t]{2mm}{\multirow{3}{*}{\rotatebox[origin=c]{90}{LSTM}}}
     		&   & 3360  &  &   &  sigmoid \\
     		&& LSTM & {\tiny ($h_{img}(t)$,} & 170 & - &  \&  \\
     		&&      & {\tiny $h_{tac}(t)$)}     &     &   &  tanh \\
    \cline{2-7}
     &\parbox[t]{2mm}{\multirow{2}{*}{\rotatebox[origin=c]{90}{FC}}}
     		& \nth{1} Output & 170 & 3          & - & - \\
     		&&                &     & {\tiny ($O^{\prime}(t)$}) &   &  \\
    \hline
    \end{tabular}
    \endgroup
    \begin{tablenotes}
        \item[1] \scriptsize {
            In and out are the number of channels for image and tactile data, and these are the number of neurons for the gate, LSTM, and FC.
            Batch normalization is applied after the n-th convolution.
            Stride and padding for the n-th convolution in image and tactile are (1, 1).
            Input is composed of an $80 \times 120$ RGB image for $I_{img}(t)$ and a $160 \times 120$ image for tactile data for $I_{tac}(t)$, and the output is a sequence of estimated object pose changes, $O^{\prime}(t) = \Delta{p^{\prime}(t)} = (\Delta{x}^{\prime}(t), \Delta{y}^{\prime}(t), \Delta{\theta}^{\prime}(t))$.
        }
    \end{tablenotes}
    \end{threeparttable}
\end{table}
\subsection{Objects}
For the target objects, we have prepared 15 objects with various size and shape (See Fig.~\ref{fig:objects}).
11 of these objects are used for training, while the remaining 4 were used to evaluate our trained network as unknown objects.
\subsection{Data Collection}
\label{sec:data_collection}
Figure~\ref{fig:hardware} shows one of the initial positions of the robot from which it starts to manipulate the object for data collection (The attached video shows more examples of the initial positions).
A table is placed in front of the robot and the object is fixed on the table with double-sided tape.
After the robot grasps the object, the gripper posture, image, and tactile data are recorded while the robot slides the object in its hand.
Since we use a parallel gripper popular in robotics, the object pose is limited to translation in the xy plane and rotating $\theta$ around the z axis.
We estimate the three DoF object pose changes given the coordinate system of the hand, which is depicted in Fig.~\ref{fig:hardware}.

The ground truth object pose change $\Delta{p}(t)$ is calculated from homogeneous transformation matrix ${^{O_1}_{O_{t}}{H}} = {{{^{R}_{O_{1}}{H}}^{-1}}{^{R}_{O_{t}}{H}}}$, where
${^{R}_{O_{t}}{H}}$ is the pose of the object at time $t$ with respect to the robot. ${^{R}_{O_{1}}{H}}$ is the pose of the object when the robot first grasps the object and makes contact.
The pose of the object is the inverse transformation of the posture of the gripper, since the object is stationary and only the gripper moves along the object. So ${^{R}_{O_{t}}{H}} = {{^{R}_{G_{t}}{H}}^{-1}}$; this can be calculated easily through forward kinematics.
 
Since we define object poses with respect to the gripper pose at initial grasp contact, what we really estimate are the relative object pose changes between the pose at the current step $t$ and initial step $1$ when the robot first makes contact with the object (when it has grasped).
Thus, the object pose change is not the absolute pose of the object.
However, the absolute pose can be calculated if the robot can estimate the absolute object pose before grasping as described in section~\ref{sec:related_works_tactile}.
Thus, the teaching signal and ground truth $O(t)$ are defined as follows:
\begin{eqnarray}
    O(t) &=& \Delta{p(t)}=(\Delta{x(t)}, \Delta{y(t)}, \Delta{\theta(t)})
    \label{eq:estiamted_value}
\end{eqnarray}

We decided to collect the teaching signals/ground truth though means of forward kinematics and by fixing the object to the table because this results in a higher accuracy as compared to other non-fixed methods such as AR marker tracking.
We also prepared the movement patterns including both translational and rotational motions, and collect data for each object (The attached video shows examples of the motions).
In order to prevent learning features from the background as the robot moves, we covered the background and desk in green clothes.
The maximum movement in translation was about 30 mm and for rotation the maximum was about 40 degrees.
For small objects like tapes, cups, scale, and wrench, only rotational movement was performed (See Fig.~\ref{fig:objects}).
For each object, the number of motions for translation, rotation, and a combination of both of them are 10, 10, and 12, respectively. 
Each grasp posture is different in each motion.
Note that even if the gripper does not occlude the object, the object can still be out of view of the camera if the object is large, and thus ``occlude`` itself.
For the trained objects in Fig.~\ref{fig:objects}, 6 out of 10 translations, 6 out of 10 rotations, and 8 out of 12 combined motions are used for training, and the remaining sets are used for evaluation.

Images, tactile, and object poses were acquired at $30~\mathrm{Hz}$, and the dataset used for training was re-sampled to $15~\mathrm{Hz}$.
The captured images were converted to gray-scale because the object pose is independent of the color of the objects.
One motion of the training dataset has a length of about 150 steps, and each step is composed of an $80 \times 120$ pixels image, a $160 \times 120$ pixels tactile image, and a 3 DoF object pose change, $(\Delta{x}(t), \Delta{y}(t), \Delta{\theta}(t))$.
\subsection{Network Design}
\label{sec:Network_Design}
The architecture of our network model is composed of two CNNs, gate, and LSTM with FC to perform DGML as shown in Fig.~\ref{fig:network_model} and as described in~\Cref{sec:proposed_method}.
We used Chainer~\cite{chainer_learningsys2015, chainermn_mlsys2017, ChainerCV2017} as deep learning library for implementation.
More details on the network parameters are shown in Table~\ref{tab:networkdesign}.
For training, we used the Huber loss as loss function $L$.
Due to computer resources, we reset the history of LSTM every 20 steps.
All our network experiments were conducted on a machine equipped with 256\,GB RAM, an Intel Xeon E5-2667v4 CPU, and eight Tesla P100-PCIE with 12GB resulting in about 24 to 48 hours of training time.
\begin{figure}[tb]
    \vspace{2mm}
    \centering
    \includegraphics[width=0.9\columnwidth]{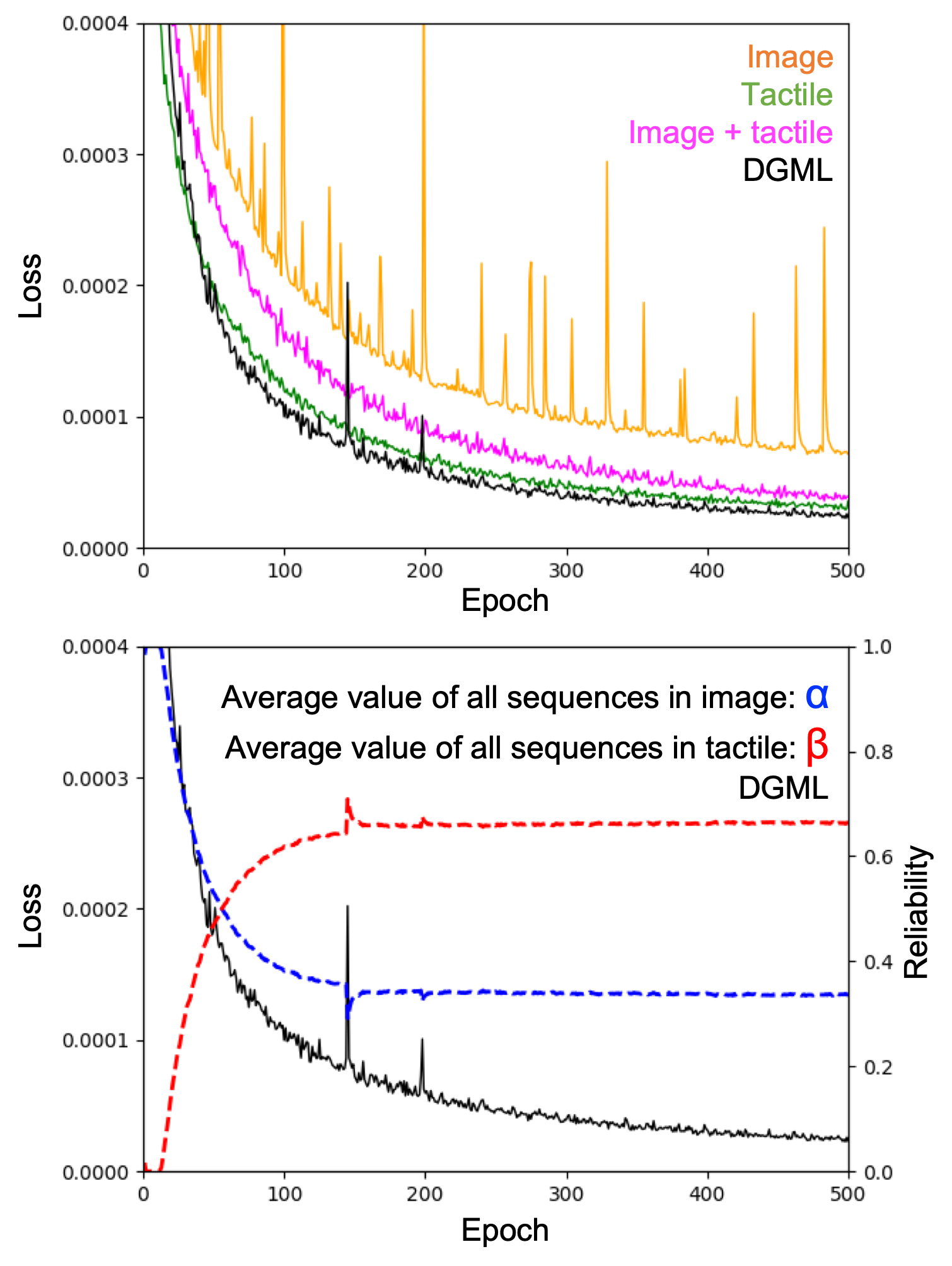}
    \caption{Learning curves and gate values of DGML.
    Note that the values of $\alpha$ and $\beta$ are the average from all training data during training per epoch.
    The values of $\alpha$ and $\beta$ are not fixed but are rather calculated from the gate through the image and tactile features at step $t$ and are inferred during run-time.}
    \label{fig:result}
\end{figure}

\section{Results}
\label{sec:result}
\subsection{Learning Curve of DGML}
As a comparison with DGML, we prepared several networks that use only images, only tactile data, and both images and tactile data with a simple connection, which is without using a gate.
The number of layers and parameters of CNN and LSTM are the same in all comparison models.
However, if one modality is disconnected, the number of input neurons of LSTM changes, since the number of input neurons is the sum of the number of neurons connected per modality.
In addition, a simple connection without using a gate is the same as DGML with fixed \emph{reliability values} $\alpha(t)=1~\&~\beta(t)=1$.

Figure~\ref{fig:result} shows the learning curves and average \emph{reliability values} of all sequences in DGML.
Note that the \emph{reliability values} of $\alpha(t)$ and $\beta(t)$ are not constants but change dynamically depending on the input $I_{img}(t)~\&~I_{tac}(t)$.
As for the \emph{reliability values} of $\alpha(t)$ for image and $\beta(t)$ for tactile, $\alpha(t)$ is greater than  $\beta(t)$ at first, but it can be seen that the \emph{reliability value} for tactile features gradually increases.
This means that the network gradually relies more on tactile data as opposed to images, which is most likely because the images become less reliable due to occlusions.
As one of the characteristics of DGML, the computational epoch until convergence is the fastest of all models since training progresses from easy-to-train modalities.

Note that the size for $I_{tac}$ is twice the size of that of $I_{img}$, but this is merely a result of the parameter search we performed, and it gave us the best performance of the network used in this study.
As a result, the size of $x_{tac}(t)$ is twice as large as $x_{img}(t)$ for the same reason.
The \emph{reliability values} of $\alpha(t)$ and $\beta(t)$ can change depending on these parameters.
However, the focus of this work is to improve the performance by introducing DGML with \emph{reliability values}, so comparing the accuracy when the same input size per modality is used is out of the scope of this work.

\begin{table}[tb]
    \vspace{2mm}
    \centering
    \caption{Inference error of in-hand object pose changes estimation}
    \label{tab:accuracy}
    \begingroup
    \scalefont{0.90}
        \begin{tabular}{c|c||c|c|c|c}
        \hline
         \multirow{2}{*}{Cond.} & \multirow{2}{*}{Model} & \multicolumn{2}{c|}{Known obj.} & \multicolumn{2}{c}{Unknown obj.} \\
         \cline{3-6}
          &  & \shortstack{Trans.\\${[}\mathrm{mm}{]}$} & \shortstack{Rot.\\${[}deg{]}$} & \shortstack{Trans.\\${[}\mathrm{mm}{]}$} & \shortstack{Rot.\\${[}deg{]}$} \\
        \hline\hline
        \multirow{4}{*}{\shortstack{Normal}}
         	& Image     & 2.10                  & 1.45                  & 3.25 & 5.47  \\
            & Tactile   & $5.99${\tiny $\times10^{-1}$}   & 1.43                  & 1.09 & 2.04  \\
            & w/o gate  & $7.48${\tiny $\times10^{-1}$}   & $8.38${\tiny $\times10^{-1}$}   & 1.01 & 1.70  \\
            & DGML      & $6.89${\tiny $\times10^{-1}$}   & $8.41${\tiny $\times10^{-1}$}   & 1.11 & 1.63  \\
        \hline
        \multirow{2}{*}{\shortstack{w/o\\image}}
            & w/o gate  & 3.00 & 8.13 & 3.01 & 9.00  \\
            & DGML      & 1.04 & 2.35 & 1.39 & 3.33  \\
        \hline
        \multirow{2}{*}{\shortstack{w/o\\tactile}}
            & w/o gate  & 5.17 & $1.35${\tiny $\times10^{1}$} & 5.02 & $1.44${\tiny $\times10^{1}$}  \\
            & DGML      & 3.49 & $1.19${\tiny $\times10^{1}$} & 3.87 & $1.39${\tiny $\times10^{1}$}  \\
        \hline
        \end{tabular}
    \endgroup
\end{table}
\begin{table}[tb]
    \centering
    \caption{The average \emph{relative value} of $\alpha$ for images and $\beta$ for tactile data}
    \label{tab:reliability_values}
    \begin{tabular}{c||c|c|c|c}
    \hline
     \multirow{2}{*}{Cond.} & \multicolumn{2}{c|}{Known obj.} & \multicolumn{2}{c}{Unknown obj.} \\
     \cline{2-5}
        & $\alpha$:image & $\beta$:tactile & $\alpha$:image & $\beta$:tactile \\
    \hline\hline
        \shortstack{with image \& tactile}     & 0.261 & 0.739  & 0.272 & 0.728  \\
    \hline
        \shortstack{w/o image}     & 0.069 & 0.931  & 0.072 & 0.928  \\
    \hline
        \shortstack{w/o tactile}   & 0.994 & 0.006  & 0.994 & 0.006  \\
    \hline
    \end{tabular}
\end{table}
\subsection{Inference Result of Object Pose}
\label{sec:inference}
Table~\ref{tab:accuracy} shows the object pose changes inference accuracy during in-hand manipulation.
The accuracy for translation and rotation is the average difference between the ground truth $(\Delta{x}(t), \Delta{y}(t), \Delta{\theta}(t))$ and inferred object pose change $(\Delta{x}^{\prime}(t), \Delta{y}^{\prime}(t), \Delta{\theta}^{\prime}(t))$ at each time step and is calculated according to equation~(\ref{eq:accuracy}):
\begin{eqnarray}
    acc_{trans} &=& \frac{1}{nT}\sum_{i=1}^{n}\sum_{t=1}^{T} (|\Delta{x(t)}-\Delta{x^{\prime}(t)}| + \nonumber\\
                & & \qquad \qquad \qquad|\Delta{y(t)}-\Delta{y^{\prime}(t)}|) \nonumber\\
    acc_{rot}   &=& \frac{1}{nT}\sum_{i=1}^{n}\sum_{t=1}^{T} (|\Delta{\theta(t)}-\Delta{\theta^{\prime}(t)}|)
    \label{eq:accuracy}
\end{eqnarray}
The ground truth is the object pose change as calculated during the data collection (see section \ref{sec:data_collection} and equation~(\ref{eq:estiamted_value})). 
We evaluated the models under normal conditions, but also by substituting either image or tactile input with random noise.
Under these conditions, we compared four different models: a model using only image input, a model using only tactile input, a model using both image and tactile (but no gate), and the proposed model.

From the results under normal conditions, the results are always better when tactile information is included as we already expected in section~\ref{sec:related_works_tactile}.
In addition, the models including tactile information can predict correctly for both known objects and unknown objects.

From the results with one of the modalities muted in Table~\ref{tab:accuracy}, we can see that the performance of the proposed DGML is the best.
When the image input is absent, the $\alpha$ value is much smaller than the $\beta$ value (See Table~\ref{tab:reliability_values}).
On the other hand, in the case of absence of tactile input, the $\beta$ value becomes almost 0.
This is because the gate decides that the \emph{reliability value} of a modality should be reduced if that modality's input is absent, resulting in near ignorance of said modality (See Table~\ref{tab:reliability_values}).
Even though the training dataset does not include data where a modality is completely absent, the gate of DGML can still deal with these situations.
\subsection{Gate Values under Noise}
\begin{figure}[t]
    \vspace{2mm}
    \centering
    \includegraphics[width=1.0\columnwidth]{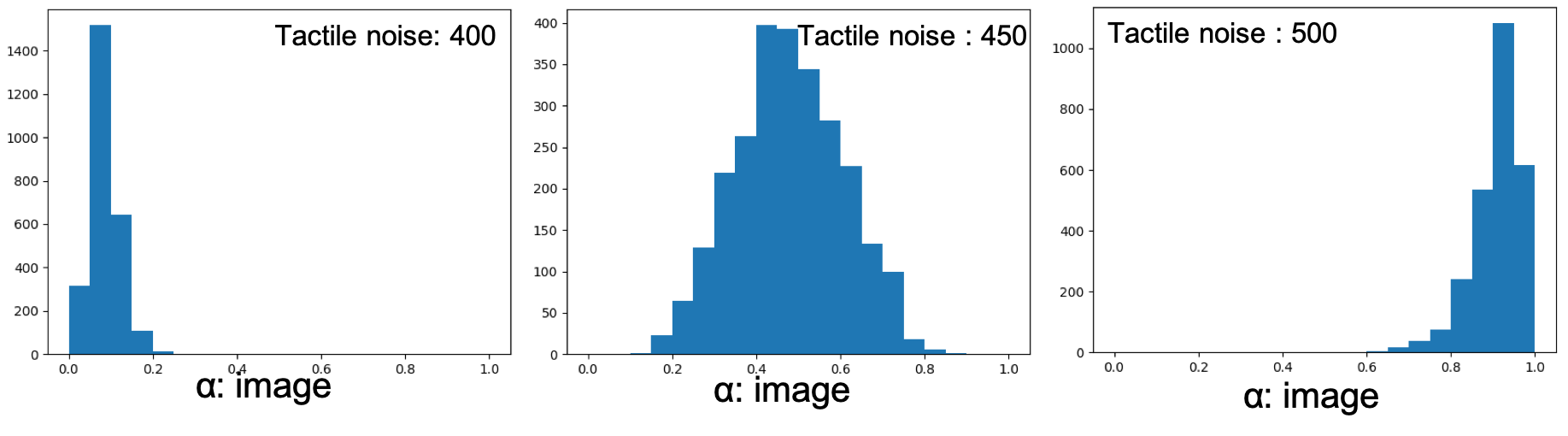}
    \caption{Histogram of image \emph{reliability values} $\alpha$ with different magnitudes of noise applied to the tactile input data}
    \label{fig:histgram_noise}
\end{figure}

In this section, we discuss the change of \emph{reliability values} when noise is applied to the input of a modality.
The noise is applied to tactile input since in-hand pose changes estimation relies more on tactile information than image information according to section~\ref{sec:inference}.
Figure~\ref{fig:histgram_noise} shows a histogram of the \emph{reliability value} $\alpha$ for images when the network infers from the original dataset, but with noise added to the tactile input $I_{tac}(t)$.
The noise is generated from a normal distribution with different variances $\sigma^2$: 400, 450, and 500 (See Fig.~\ref{fig:histgram_noise}).

From Fig.~\ref{fig:histgram_noise}, it can be seen that the \emph{reliability value} for images increases as the noise to the tactile input increases.
The gate can correctly recognize the noise applied to the tactile input data.
The strength of using DGML is that DGML helps to understand the network behavior through the \emph{reliability values} of the gate because the gate represents how much each modality should be used.
Furthermore, if a sensor is broken, the \emph{reliability value} for that sensor is always close to 0, thus DGML can recognize sensor failures.

\begin{figure}[t]
    \vspace{2mm}
    \centering
    \includegraphics[width=0.85\columnwidth]{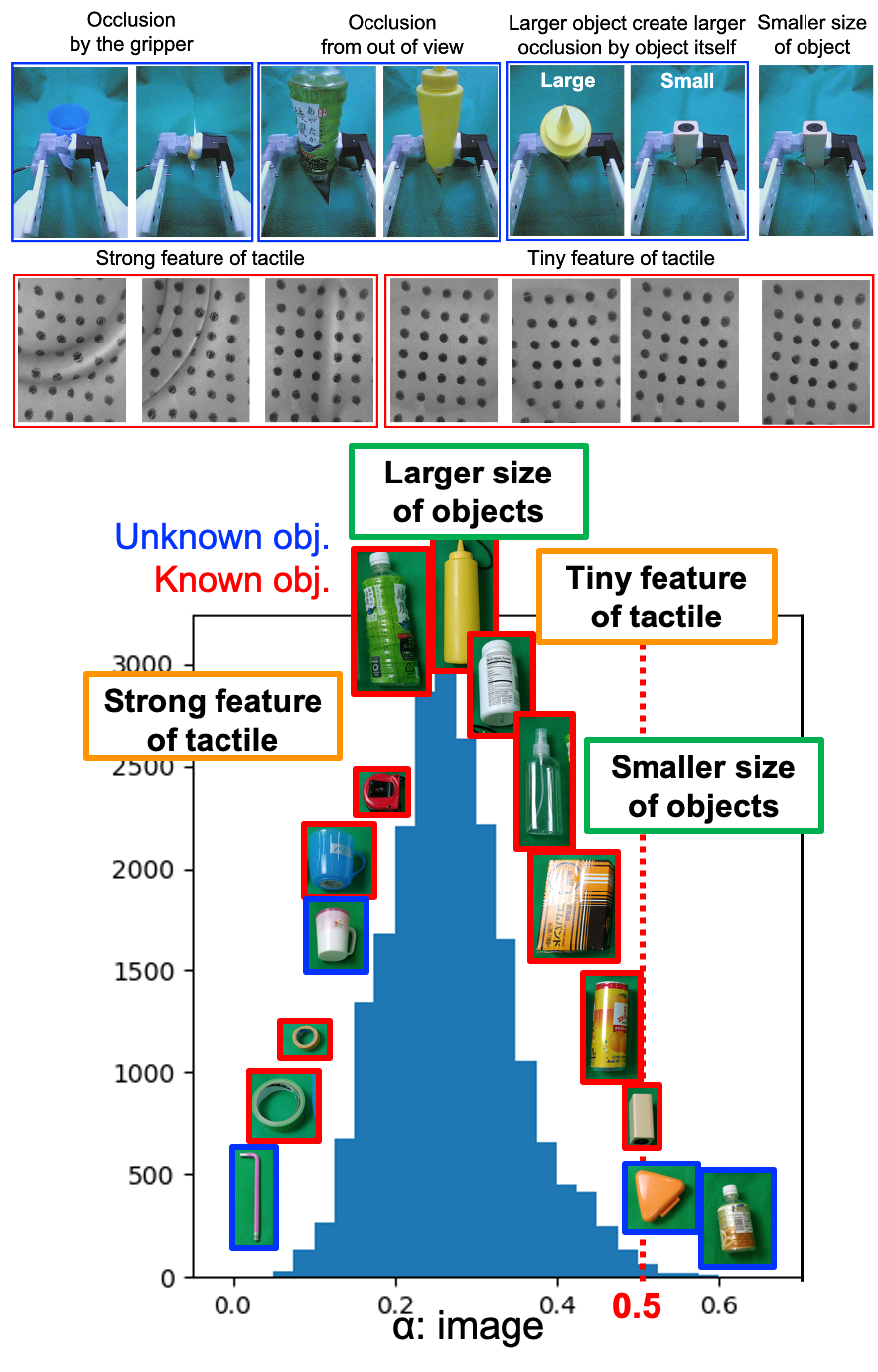}
    \caption{Histogram of image reliability values $\alpha$.
    The objects surrounded by a red color frame are known objects, whereas the objects surrounded by a blue color frame are unknown objects.
    The size of the object in each image is the same as the ratio of the actual size.}
    \label{fig:histgram}
\end{figure}
\subsection{Reliability Values of Objects}
In this section, we discuss the change of the \emph{reliability values} of objects due to object shape, size, and occlusions.
We visualize the \emph{reliability value} of images, $\alpha$, by creating a histogram with all the inferred values of $\alpha$ through untrained motions with both known and unknown objects (See Fig.~\ref{fig:histgram}).
Most of the reliability values of $\alpha$ are smaller than 0.5, which indicates that the network relies more on the tactile features in those cases.
From the comparison between only using images and only using tactile data in section~\ref{sec:inference}, we draw the hypothesis that the \emph{reliability values} for images are low because the method that uses only tactile data has higher accuracy than the one using only images (See Table~\ref{tab:accuracy}), which seems to indicate that tactile data is more useful for this task.

The objects are depicted on the most frequently occurring $\alpha$ on the histogram in Fig.~\ref{fig:histgram}.
The ratio between the object sizes in the image and the actual object sizes is the same.
Since objects with $\alpha$ values between 0 and 0.2 such as tapes, wrenches, and cups tend to provide rich tactile information due to their unique shape and/or surface, the network relies more on tactile information for these objects.
Conversely, image information is reliable for objects with no surface irregularities and are displayed with $\alpha$ values between 0.4 and 0.6 such as boxes and cans, for which tactile features are difficult to observe from.
Although some objects have no tactile features on the surface, the reliability of the image is low when such an object is large and thus occludes itself or doesn't fit within the camera view, as opposed to small objects (See top row of Fig.~\ref{fig:histgram}).
For example, the wood, the plastic coffee bottle, and the orange triangle are small, thus their movements can be observed from images.
In contrast, large objects such as the mustard and green plastic bottles create occlusions, which reduces the reliability of image information, therefore the network uses more tactile information.
Therefore, we can say that the proposed method is effective in determining the \emph{reliability value} of each modality according to the object shape, size, and occlusions.

\section{Conclusion}
\label{sec:conclusion}
In this paper, we proposed a method for in-hand object pose changes estimation using image and tactile data, while also predicting the modalities' \emph{reliability values}, called deep gated multi-modal learning (DGML).
The proposed method can estimate not only known object pose changes but also unknown object pose changes during grasping, since it doesn't need 3D models.
Moreover, the modalities' \emph{reliability values} can be changed dynamically and automatically by the network depending on situations such as sensor failure and different magnitudes of noise.
Visualization of \emph{reliability values} helps to understand the network behavior such as which modality should be utilized and how much.
Furthermore, by using the proposed method, computation efficiency for training has been improved.

For future work, we will develop a system of in-hand object manipulation to integrate the object pose estimator.

\section*{ACKNOWLEDGMENT} \small
The authors would like to thank Wilson Ko for helping to discuss and writting, and Koichi Nishiwaki and Tianyi Ko for proofreading.
\bibliographystyle{IEEEtran} 
\bibliography{bibliography}

\begin{thebibliography}{10}
\providecommand{\url}[1]{#1}
\csname url@rmstyle\endcsname
\providecommand{\newblock}{\relax}
\providecommand{\bibinfo}[2]{#2}
\providecommand\BIBentrySTDinterwordspacing{\spaceskip=0pt\relax}
\providecommand\BIBentryALTinterwordstretchfactor{4}
\providecommand\BIBentryALTinterwordspacing{\spaceskip=\fontdimen2\font plus
\BIBentryALTinterwordstretchfactor\fontdimen3\font minus
  \fontdimen4\font\relax}
\providecommand\BIBforeignlanguage[2]{{%
\expandafter\ifx\csname l@#1\endcsname\relax
\typeout{** WARNING: IEEEtran.bst: No hyphenation pattern has been}%
\typeout{** loaded for the language `#1'. Using the pattern for}%
\typeout{** the default language instead.}%
\else
\language=\csname l@#1\endcsname
\fi
#2}}

\bibitem{yousef2011tactile}
H.~Yousef, \emph{et~al.}, ``Tactile sensing for dexterous in-hand manipulation
  in robotics—a review,'' \emph{Sensors and Actuators A: Physical}, vol. 167,
  no.~2, pp. 171 -- 187, 2011, solid-State Sensors, Actuators and Microsystems
  Workshop.

\bibitem{nikhil2018inhand}
C.-D. Nikhil, \emph{et~al.}, ``In-hand manipulation via motion cones,''
  \emph{Robotics: Science and Systems (RSS)}, 2018.

\bibitem{choi2012voting}
C.~Choi, \emph{et~al.}, ``Voting-based pose estimation for robotic assembly
  using a 3d sensor,'' in \emph{2012 IEEE International Conference on Robotics
  and Automation}, May 2012, pp. 1724--1731.

\bibitem{choi20123d}
C.~Choi and H.~I. Christensen, ``3d pose estimation of daily objects using an
  rgb-d camera,'' in \emph{2012 IEEE/RSJ International Conference on
  Intelligent Robots and Systems}, Oct 2012, pp. 3342--3349.

\bibitem{aldoma2012tutorial}
A.~Aldoma, \emph{et~al.}, ``Tutorial: Point cloud library: Three-dimensional
  object recognition and 6 dof pose estimation,'' \emph{IEEE Robotics
  Automation Magazine}, vol.~19, no.~3, pp. 80--91, Sept 2012.

\bibitem{ICP}
S.~Rusinkiewicz and M.~Levoy, ``Efficient variants of the icp algorithm,'' in
  \emph{Proceedings Third International Conference on 3-D Digital Imaging and
  Modeling}, May 2001, pp. 145--152.

\bibitem{krull2015learning}
A.~Krull, \emph{et~al.}, \emph{\BIBforeignlanguage{English}{Learning
  Analysis-by-Synthesis for 6D Pose Estimation in RGB-D Images}}, 12 2015, pp.
  954--962.

\bibitem{xiang2017posecnn}
Y.~Xiang, \emph{et~al.}, ``Posecnn: A convolutional neural network for 6d
  object pose estimation in cluttered scenes,'' \emph{Robotics: Science and
  Systems (RSS)}, 2018.

\bibitem{Wang2019DenseFusion}
C.~Wang, \emph{et~al.}, ``Densefusion: 6d object pose estimation by iterative
  dense fusion,'' in \emph{Proceedings of the IEEE Conference on Computer
  Vision and Pattern Recognition}, 2019, pp. 3343--3352.

\bibitem{schwarz2015rgbd}
M.~Schwarz, \emph{et~al.}, ``Rgb-d object recognition and pose estimation based
  on pre-trained convolutional neural network features,'' in \emph{2015 IEEE
  International Conference on Robotics and Automation (ICRA)}, May 2015, pp.
  1329--1335.

\bibitem{hodan2018bop}
T.~Hodan, \emph{et~al.}, ``Bop: Benchmark for 6d object pose estimation,'' in
  \emph{Proceedings of the European Conference on Computer Vision (ECCV)},
  2018, pp. 19--34.

\bibitem{hu2019segmentation}
Y.~Hu, \emph{et~al.}, ``Segmentation-driven 6d object pose estimation,'' in
  \emph{Proceedings of the IEEE Conference on Computer Vision and Pattern
  Recognition}, 2019, pp. 3385--3394.

\bibitem{andrychowicz2020learning}
O.~M. Andrychowicz, \emph{et~al.}, ``Learning dexterous in-hand manipulation,''
  \emph{The International Journal of Robotics Research}, vol.~39, no.~1, pp.
  3--20, 2020.

\bibitem{tomo2016uskin}
T.~P. Tomo, \emph{et~al.}, ``A modular, distributed, soft, 3-axis sensor system
  for robot hands,'' in \emph{2016 IEEE-RAS 16th International Conference on
  Humanoid Robots (Humanoids)}, Nov 2016, pp. 454--460.

\bibitem{tomo2018uskin}
T.~O. Tomo, \emph{et~al.}, ``Covering a robot fingertip with uskin: A soft
  electronic skin with distributed 3-axis force sensitive elements for robot
  hands,'' \emph{IEEE Robotics and Automation Letters}, vol.~3, no.~1, pp.
  124--131, 2017.

\bibitem{yuan2017gelsight}
W.~Yuan, \emph{et~al.}, ``{GelSight: High-Resolution Robot Tactile Sensors for
  Estimating Geometry and Force},'' \emph{Sensors}, vol.~17, no.~12, p. 2762,
  2017.

\bibitem{dong2017improved}
S.~Dong, \emph{et~al.}, ``Improved gelsight tactile sensor for measuring
  geometry and slip,'' in \emph{2017 IEEE/RSJ International Conference on
  Intelligent Robots and Systems (IROS)}.\hskip 1em plus 0.5em minus
  0.4em\relax IEEE, 2017, pp. 137--144.

\bibitem{ohmura2006conformable}
Y.~Ohmura, \emph{et~al.}, ``{Conformable and Scalable Tactile Sensor Skin for
  Curved Surfaces},'' in \emph{IEEE International Conference on Robotics and
  Automation (ICRA)}, 2006, pp. 1348--1353.

\bibitem{iwata2009design}
H.~Iwata and S.~Sugano, ``{Design of Human Symbiotic Robot TWENDY-ONE},'' in
  \emph{IEEE International Conference on Robotics and Automation (ICRA)}, 2009,
  pp. 580--586.

\bibitem{mittendorfer2011humanoid}
P.~Mittendorfer and G.~Cheng, ``{Humanoid Multimodal Tactile-sensing
  Modules},'' \emph{IEEE Transactions on robotics}, vol.~27, no.~3, pp.
  401--410, 2011.

\bibitem{fishel2012sensing}
J.~A. Fishel and G.~E. Loeb, ``{Sensing Tactile Microvibrations with the
  BioTac—Comparison with Human Sensitivity},'' in \emph{IEEE RAS \& EMBS
  International Conference on Biomedical Robotics and Biomechatronics
  (BioRob)}, 2012, pp. 1122--1127.

\bibitem{paulino2017low}
T.~Paulino, \emph{et~al.}, ``{Low-cost 3-axis Soft Tactile Sensors for the
  Human-Friendly Robot Vizzy},'' in \emph{IEEE International Conference on
  Robotics and Automation (ICRA)}, 2017, pp. 966--971.

\bibitem{johnson2009retrographic}
M.~K. Johnson and E.~H. Adelson, ``{Retrographic Sensing for the Measurement of
  Surface Texture and Shape},'' in \emph{IEEE Conference on Computer Vision and
  Pattern Recognition (CVPR)}, 2009, pp. 1070--1077.

\bibitem{Calandra2017}
R.~Calandra, \emph{et~al.}, ``{More Than a Feeling : Learning to Grasp and
  Regrasp using Vision and Touch},'' no. Nips, pp. 1--10, 2017.

\bibitem{bimbo2015global}
J.~Bimbo, \emph{et~al.}, ``Global estimation of an object’s pose using
  tactile sensing,'' \emph{Advanced Robotics}, vol.~29, no.~5, pp. 363--374,
  2015.

\bibitem{gao2016deep}
Y.~Gao, \emph{et~al.}, ``Deep learning for tactile understanding from visual
  and haptic data,'' in \emph{2016 IEEE International Conference on Robotics
  and Automation (ICRA)}.\hskip 1em plus 0.5em minus 0.4em\relax IEEE, 2016,
  pp. 536--543.

\bibitem{Bimbo2012}
J.~Bimbo, \emph{et~al.}, ``Object pose estimation and tracking by fusing visual
  and tactile information,'' in \emph{2012 IEEE International Conference on
  Multisensor Fusion and Integration for Intelligent Systems (MFI)}, Sept 2012,
  pp. 65--70.

\bibitem{honda1998real}
K.~Honda, \emph{et~al.}, ``Real-time pose estimation of an object manipulated
  by multi-fingered hand using 3d stereo vision and tactile sensing,'' in
  \emph{Proceedings. 1998 IEEE/RSJ International Conference on Intelligent
  Robots and Systems. Innovations in Theory, Practice and Applications (Cat.
  No. 98CH36190)}, vol.~3.\hskip 1em plus 0.5em minus 0.4em\relax IEEE, 1998,
  pp. 1814--1819.

\bibitem{hebert2011fusion}
P.~Hebert, \emph{et~al.}, ``Fusion of stereo vision, force-torque, and joint
  sensors for estimation of in-hand object location,'' in \emph{2011 IEEE
  International Conference on Robotics and Automation}.\hskip 1em plus 0.5em
  minus 0.4em\relax IEEE, 2011, pp. 5935--5941.

\bibitem{Bimbo2016}
J.~Bimbo, \emph{et~al.}, ``In-hand object pose estimation using
  covariance-based tactile to geometry matching,'' \emph{IEEE Robotics and
  Automation Letters}, vol.~1, no.~1, pp. 570--577, Jan 2016.

\bibitem{takahashi2017dynamic}
K.~Takahashi, \emph{et~al.}, ``Dynamic motion learning for multi-dof
  flexible-joint robots using active--passive motor babbling through deep
  learning,'' \emph{Advanced Robotics}, vol.~31, no.~18, pp. 1002--1015, 2017.

\bibitem{lee2019making}
M.~A. Lee, \emph{et~al.}, ``Making sense of vision and touch: Self-supervised
  learning of multimodal representations for contact-rich tasks,'' in
  \emph{2019 International Conference on Robotics and Automation (ICRA)}.\hskip
  1em plus 0.5em minus 0.4em\relax IEEE, 2019, pp. 8943--8950.

\bibitem{hatori2018interactive}
J.~Hatori, \emph{et~al.}, ``Interactively picking real-world objects with
  unconstrained spoken language instructions,'' \emph{2018 IEEE International
  Conference on Robotics and Automation (ICRA)}, 2018.

\bibitem{noda2014multimodal}
K.~Noda, \emph{et~al.}, ``Multimodal integration learning of robot behavior
  using deep neural networks,'' \emph{Robotics and Autonomous Systems},
  vol.~62, no.~6, pp. 721--736, 2014.

\bibitem{kim2018robust}
J.~Kim, \emph{et~al.}, ``Robust deep multi-modal learning based on gated
  information fusion network,'' \emph{arXiv preprint arXiv:1807.06233}, 2018.

\bibitem{hu2018squeeze}
J.~Hu, \emph{et~al.}, ``Squeeze-and-excitation networks,'' in \emph{Proceedings
  of the IEEE conference on computer vision and pattern recognition}, 2018, pp.
  7132--7141.

\bibitem{arevalo2017gated}
J.~Arevalo, \emph{et~al.}, ``Gated multimodal units for information fusion,''
  \emph{arXiv preprint arXiv:1702.01992}, 2017.

\bibitem{chainer_learningsys2015}
S.~Tokui, \emph{et~al.}, ``Chainer: a next-generation open source framework for
  deep learning,'' in \emph{Proceedings of Workshop on Machine Learning Systems
  (LearningSys) in The Twenty-ninth Annual Conference on Neural Information
  Processing Systems (NIPS)}, 2015.

\bibitem{chainermn_mlsys2017}
A.~Takuya, \emph{et~al.}, ``{ChainerMN: Scalable Distributed Deep Learning
  Framework},'' in \emph{Proceedings of Workshop on ML Systems in The
  Thirty-first Annual Conference on Neural Information Processing Systems
  (NIPS)}, 2017.

\bibitem{ChainerCV2017}
Y.~Niitani, \emph{et~al.}, ``Chainercv: a library for deep learning in computer
  vision,'' in \emph{ACM Multimedia}, 2017.

\end{thebibliography}
\end{document}